\pdfoutput=1
\documentclass[letterpaper]{article} %
\usepackage{aaai2026}  %
\usepackage{times}  %
\usepackage{helvet}  %
\usepackage{courier}  %
\usepackage[hyphens]{url}  %
\usepackage{graphicx} %
\usepackage{natbib}  %
\usepackage{caption} %
\usepackage{algorithm}

\usepackage{algpseudocode}
\usepackage{newfloat}
\usepackage{listings}

\usepackage{graphicx}
\usepackage{booktabs} %
\usepackage{url}
\usepackage{amsmath}
\usepackage{subcaption}
\usepackage{pgfplots}
\pgfplotsset{compat=1.18} 
\usepgfplotslibrary{groupplots, fillbetween, statistics}
\usepackage{amsfonts}
\usepackage{amsthm}
\usepackage{booktabs, multirow, xcolor, colortbl}
\usepackage{array}
\usepackage{caption}
\usepackage{tikz}
\usepackage{placeins}
\usepackage{tikz}
\usepackage{float}
\usepackage{mathtools}
\usetikzlibrary{positioning,arrows.meta,fit,backgrounds,calc}

\usepackage{amsmath}
\usepackage{amssymb}
\usepackage{mathtools}
\usepackage{amsthm}

\usepackage[capitalize,noabbrev]{cleveref}

\theoremstyle{plain}
\newtheorem{theorem}{Theorem}[section]
\newtheorem{proposition}[theorem]{Proposition}

\theoremstyle{definition}

\theoremstyle{remark}

\usepackage[textsize=tiny]{todonotes}

\newcommand{\subfour}[1]{\vspace*{3mm}{\noindent\bf #1}}

\DeclareCaptionStyle{ruled}{labelfont=normalfont,labelsep=colon,strut=off} %
\floatstyle{ruled}
\newfloat{listing}{tb}{lst}{}
\floatname{listing}{Listing}

\title{	Beyond Raw Similarity: Prior-Aware Fusion for Out-of-Distribution Detection}
\author{
    Lakpa Tamang \textsuperscript{\rm 1} \textsuperscript{\rm ,} \textsuperscript{\rm 2},
    Mohamed Reda Bouadjenek \textsuperscript{\rm 1},
    Saman Forouzandeh \textsuperscript{\rm 2},
    Kamal Berahmand \textsuperscript{\rm 2},
    Mahdi Jalili \textsuperscript{\rm 2},
    Sunil Aryal \textsuperscript{\rm 1}
}
\affiliations{
    \textsuperscript{\rm 1}School of Information Technology, Deakin University\\
Geelong, Australia\\
    \textsuperscript{\rm 2}Department of Electrical and Electronics Engineering, RMIT University\\
        Melbourne, Australia\\
    \{lakpa.tamang, saman.forouzandeh, kamal.berahmand, mahdi.jalili\}@rmit.edu.au \\

        \{reda.bouadjenek, sunil.aryal\}@deakin.edu.au
}

\usepackage{bibentry}

\begin{document}

\maketitle

\begin{abstract}
Out-of-distribution (OOD) detection in vision--language models presents
unique challenges due to the reliance on cross-modal similarity. Even when
embeddings are well separated for in-distribution (ID) data, similarity scores
can remain overconfident for OOD inputs. A common remedy is to adapt the model on
ID data, but adaptation discards the pretrained geometry that made the
similarity meaningful. We propose multi-modal logit fusion (\textbf{MM-Log}), which keeps both by pairing a
gradient-free zero-shot prior with a task-adapted branch in each modality
and fuses their four cross-modal similarity channels into a single OOD score.
Because the zero-shot priors are never adapted, the cross-channels act as a
consistency check between the pretrained and adapted geometries -- an OOD input
that fools the adapted branch through a shortcut must additionally agree with
frozen CLIP to retain a high score. We formalize this as a margin bound showing
that the fused score is provably penalized by zero-shot--adapted disagreement, a
signal to which adapted-only scores are structurally blind. MM-Log requires no
auxiliary OOD data, negative prompts, or test-time adaptation, and trains on
few-shot ID data alone. Across Far-OOD, Near-OOD, and Hard-OOD benchmarks MM-Log
consistently outperforms strong CLIP-based baselines.
\end{abstract}

\section{Introduction}

Modern machine learning models are inherently vulnerable to inputs that fall outside their training distribution. When confronted with unseen classes, novel concepts, or shifted data conditions, these models often extrapolate with unjustified confidence, leading to erroneous and potentially harmful decisions \cite{tamang2025handling}. To address this, out-of-distribution (OOD) detection seeks to determine whether a given test input originates from the model’s in-distribution (ID) data or represents an unknown case. Traditionally, OOD detection has been studied primarily in single-modality settings \cite{hendrycks2018deep, liu2020energy}, leveraging confidence scores or feature statistics to detect anomalous inputs. However, recent advancements in Vision–language models (VLMs) such as CLIP \cite{radford2021learning} have substantially expanded this scope as they offer strong zero-shot transfer learning, naturally operate on both image and text inputs, and produce a shared embedding space where cross-modal similarity can be used for downstream tasks. 

Early CLIP-based OOD detection methods such as Maximum Concept Matching (MCM) \cite{ming2022delving}, and Global-local MCM \cite{miyai2023zero} primarily rely on the model’s inherent cross-modal similarity geometries. While conceptually appealing, their detection performance is limited in practice, indicating that the raw similarity structure of pretrained VLMs is often insufficient for reliable OOD separation. Other subsequent methods move beyond the pretrained model’s natural similarity geometry. One of the common techniques is to introduce auxiliary OOD signals \cite{esmaeilpour2022zero, miao2025auxiliary}, synthetic negatives \cite{jiang2024negative, zhang2024lapt}, or surrogate distributions \cite{miyai2023locoop, yu2024self} to calibrate scores and shape the decision boundary. Another popular strategy is negative prompting \cite{wang2023clipn}, where the detector queries the model with carefully designed “non-class” or “none-of-the-above” prompts to elicit low similarity for OOD inputs. These techniques can be effective, but also complicate the pipeline as they introduce prompt sensitivity, increase dependency on curated or generated negatives, and require additional hyperparameters and heuristics for processing surrogate data which may not transfer across datasets, modalities, or deployment conditions. 

In this paper, we argue that a key failure mode in multi-modal OOD detection is not merely the absence of ``OOD knowledge'', but rather the miscalibration and brittleness of cross-modal similarity under distribution shift. While strong image–text alignment enables VLMs to learn discriminative embeddings across diverse ID categories, their similarity scores remain susceptible to overconfidence induced by spurious correlations, shortcut cues, and modality-specific artifacts. \cite{yang2025escaping}. This indicates that strong cross-modal agreement alone, while sufficient for classification, is not enough for reliable OOD detection. We therefore hypothesize that explicitly strengthening and constraining semantic agreement between modalities for ID data can make OOD detection substantially easier. This is because OOD samples tend to violate such an agreement in subtle ways, through novel concepts, mismatched descriptions, or shifts that perturb one modality while leaving the other relatively unchanged. Consequently, by enforcing tighter cross-modal consistency, even small departures from the ID manifold can be amplified and detected as OOD. With this motivation, we propose MM-Log (\textbf{M}ulti-\textbf{M}odal \textbf{Log}it Fusion), a simple cross-modal OOD detection approach built around two concepts: (i) enforce tighter alignment between ID image and text embeddings in the shared representation space, and (ii) preserve the rich semantic knowledge of pretrained VLMs by using zero-shot representations as a regularizing prior. Specifically, instead of relying on auxiliary OOD signals or handcrafted negative prompts, we adapt the VLM solely with ID data, learning an embedding geometry that pulls matched image–text pairs closer together and pushes mismatched pairs farther apart. This creates an ID manifold with higher cross-modal coherence and a similarity landscape that is less permissive to OOD inputs.

Our key contributions are summarized as follows:
\begin{enumerate}
\item We propose \textbf{MM-Log}, which pairs a gradient-free zero-shot prior
with a task-adapted branch in each modality and fuses their four cosine
channels into a single OOD score, trained on
few-shot ID data alone without auxiliary outliers, negative prompts, or
test-time adaptation.

\item We show that the fused score admits an exact factorization whose magnitude
is governed by the angular disagreement between the pretrained and adapted
geometries, and prove a margin bound (Prop.~\ref{prop:margin}) quantifying how
this disagreement penalizes the score; adapted-only scores are structurally blind
to this signal.

\item We validate MM-Log across Far-, Near-, and Hard-OOD benchmarks, where it
outperforms strong CLIP-based baselines, and isolate the contribution of each
cross-modal interaction through a channel-wise ablation.
\end{enumerate}

\section{Related Works}

\subfour{Single-modal OOD Detection Methods}
 OOD detection has been extensively studied in single-modal settings, particularly in image classification through an image encoder such as ResNet. Early approaches on single-modal OOD detection rely on the predictive confidence of discriminative classifiers, such as maximum softmax probability (MSP)~\cite{hendrycks2016baseline} and its extensions, as well as energy-based scores that reinterpret classifier logits as unnormalized densities~\cite{liu2020energy}. Other lines of work leverage feature-space statistics, including distance-based methods such as the Mahalanobis distance~\cite{lee2018simple}, or model data likelihoods estimated by generative models~\cite{serra2019input}. Beyond post-hoc scoring functions, several methods explicitly incorporate OOD information as auxiliary outliers during training \cite{hendrycks2018deep,tamang2024margin, ming2022poem, chen2021atom}. In contrast to these single-modal approaches, our work targets OOD detection in large VLMs, where the presence of a shared cross-modal embedding space introduces both new challenges and opportunities that are not captured by traditional unimodal formulations.

\subsection{Vision Language Model-based OOD Detection}
With the emergence of large-scale VLMs, recent work has started to investigate OOD detection in multimodal and cross-modal settings by exploiting the shared embedding space learned by models such as CLIP. Early VLM-based OOD detection approaches primarily operate directly on CLIP-style image--text similarity scores. Representative methods, including MCM~\cite{ming2022delving} and its extension GL-MCM~\cite{miyai2025gl}, use the maximum similarity between an image and a set of class prompts as an OOD confidence score, based on the intuition that ID samples should exhibit strong alignment with at least one semantic class. Another prominent line of research focuses on prompt engineering and prompt learning, where carefully designed or learned prompts such as negative or ``none-of-the-above'' descriptions~\cite{jiang2024negative, zhang2024lapt, wang2023clipn} are introduced to induce lower similarity responses for OOD inputs. A complementary direction incorporates auxiliary OOD data or synthetically generated outliers~\cite{miyai2023locoop, yu2024self, miao2025auxiliary} to explicitly shape the decision boundary and improve score calibration. Despite their empirical success, these methods often rely heavily on the availability and quality of proxy OOD information, as well as the careful construction of negative labels, which limits their robustness in realistic and ubiquitous deployment settings. Moreover, they tend to be sensitive to prompt design choices and frequently introduce additional hyperparameters that may not generalize well across datasets or modalities. In contrast, our approach adopts a minimal and principled setting: we rely exclusively on ID data while avoiding auxiliary OOD supervision and negative prompting. Rather than introducing additional signals, we directly improve the cross-modal similarity structure by tightening image--text alignment for ID pairs.

\section{Preliminaries}
\subfour{OOD Detection Setup: }Let $\mathbf{x}\in\mathcal{X}_{train}$ and $y\in\mathcal{Y}_{train} = \{1,2,\ldots, C\}$ represent a set of input images and their labels, respectively, drawn from the ID dataset $\mathbb{P}_\text{ID}$ with $C$ classes. Subsequently, for a test dataset $\mathcal{X}_{test}$ drawn from OOD dataset $\mathbb{P}_\text{OOD}$, it holds that $\mathbb{P}_\text{ID} \cap \mathbb{P}_\text{OOD} = \phi$. Then, formally the objective of OOD detection during testing remain two-folds: (i) for $\mathcal{X}_{test} \sim \mathbb{P}_\text{ID}$, classify it into correct ID class; (ii) for $\mathcal{X}_{test} \sim \mathbb{P}_\text{OOD}$, reject it as OOD data. 
Formally, the second objective can be achieved through a simple scoring mechanism:
\begin{equation} \label{eq:2}
  G(\mathbf{x})=\begin{cases}
    \text{ID}, & \text{if $s(\mathbf{x})\geq\gamma$}.\\
    \text{OOD}, & \text{otherwise}.
  \end{cases}
\end{equation}
where, $G(\cdot)$ is an OOD detector, $s(\mathbf{x})$ denotes the confidence score of $\mathbf{x} \in \mathcal{X}_{test}$, and $\gamma \in \mathbb{R}$ is a threshold computed based on some heuristics.

\subfour{Revisiting CLIP and Prompt Learning: }With prompt learning, a simple natural-language text prompt is transformed into a set of learnable prompt vectors that are optimized to encode task-specific semantics, enabling tighter and more effective alignment with the model’s visual representations. For each class $c$, consider a manually designed textual prompt as
\[
\mathbf{p} = \text{``a photo of a } <y_c>\text{''}.
\]
Following CoOp~\cite{zhou2022conditional}, prompt learning reformulates the textual prompt as a sequence of learnable context tokens,
$\mathbf{u} = \{\mathbf{v}_1, \mathbf{v}_2, \ldots, \mathbf{v}_N, \mathbf{w}_c\}$, where $\mathbf{w}_c$ denotes the word embedding of class $c$, and $\{\mathbf{v}_n\}_{n=1}^N$ are $N$ learnable context vectors shared across all classes. We distinguish between two versions of this class prompt. The initialization prompt, $\mathbf{u}^{(0)}$, is constructed from the handcrafted template "a photo of a $<y_c>$" and kept frozen throughout training, preserving the pretrained VLM's zero-shot semantics. The adapted prompt, $\mathbf{u}^{(a)}$, shares the same initialization but is optimized via gradient descent, allowing it to capture task-specific adaptations from the ID training data. Then, given a pretrained CLIP model, the image and text embeddings are obtained as $\mathbf{z} = f(\mathbf{x}), \mathbf{t} = g(\mathbf{u})$,
where $f(\cdot)$ and $g(\cdot)$ denote the image encoder and text encoder, respectively, and the embeddings are $L_2$-normalized. The cross-modal similarity is then computed via cosine similarity. For downstream tasks such as image recognition, the probability of assigning image $\mathbf{x}$ to class $c$ is then given by
\begin{equation}
p(y = c \mid \mathbf{x}) =
\frac{
e^{S^c/ \tau
}}{
\sum_{i=1}^{C}
e^{S^i/\tau}
},
\end{equation}
where $\tau$ is a softmax temperature parameter, and $S^c = \mathbf{z}^\top\cdot\mathbf{t}$ represents cosine similarity between the image features and its corresponding textual prompt vector for class $c$.

\section{MM-Log: Multi-Modal Logit Fusion}
\begin{figure*}[ht!]
\centering
\includegraphics[width=0.9\textwidth]{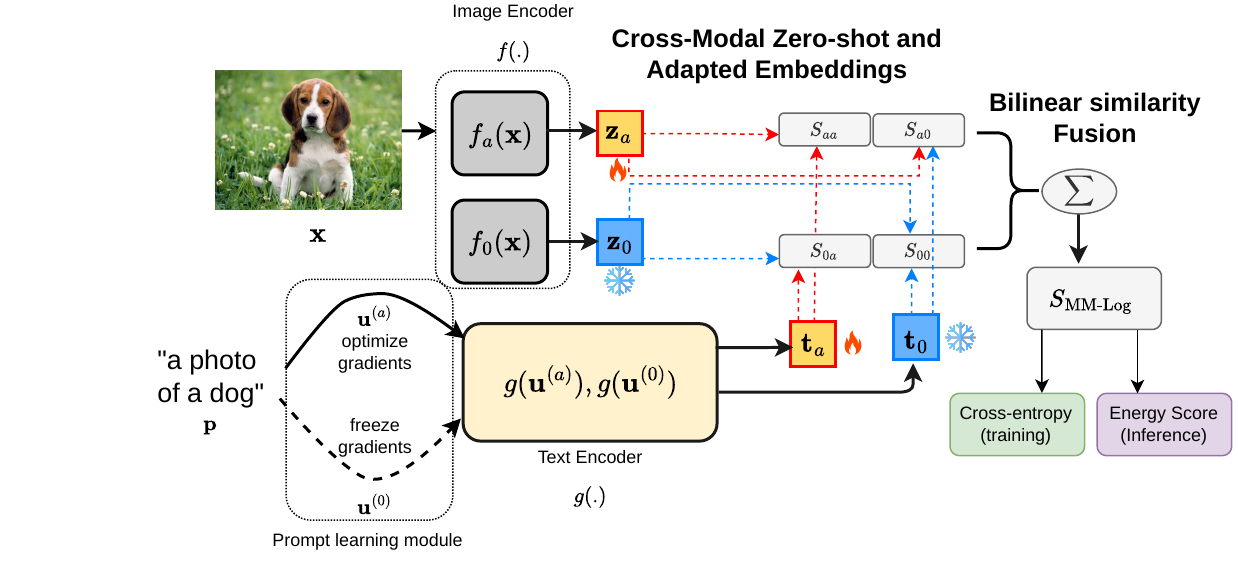}

\caption{Schematic overview of MM-Log. The embedding outcomes are \emph{zero-shot} feature ($\mathbf{z}_0,\mathbf{t}_0$) and an \emph{adapted} feature ($\mathbf{z}_a,\mathbf{t}_a$). On the image side these come from a frozen encoder $f_0$ and a encoder adapter $f_a$. On the text side a \emph{single} shared prompt learner are evaluated at two prompt states: the template-initialized $\mathbf{u}^{(0)}$ (cached, no gradient) gives $\mathbf{t}_0$, and the trained prompt $\mathbf{u}_a$ gives $\mathbf{t}_a$. The four zero-shot/adapted cosine pairings form the channels $S_{00},S_{0a},S_{a0},S_{aa}$, summed into $S_{\text{MM-Log}}$; the fused score is supervised by cross-entropy during training, and its free energy serves as the OOD score at inference.}
\label{fig:overview}
\end{figure*}

We propose \textbf{MM-Log} (Multi-Modal Logit Fusion), that jointly leverages (i) intrinsic zero-shot CLIP features and (ii) task-adapted features from the few-shot ID samples of the downstream dataset. MM-Log integrates these complementary signals through a simple bilinear fusion mechanism of the logits derived from their respective multi-modal similarity function. The outcome is a model that substantially enhances both ID accuracy and OOD separability without requiring external OOD data. The schematic representation of MM-Log's working procedure is shown in Figure \ref{fig:overview}. The MM-Log model consists of two components: (i) Cross-Modal Zero-Shot and Adapted Embeddings, and (ii) Bilinear Similarity Fusion, both of which are discussed in the following sections.

\subfour{Cross-Modal Zero-Shot and Adapted Embeddings: }
\label{sec:zeroshot-adapted}
We begin by extracting embeddings from the vanilla CLIP and simultaneously adapting to the downstream dataset on-the-fly during training. In particular, we aim in generating zero-shot and adapted embeddings for each modality. The source of MM-Log's zero-shot text prior, lies in when the prompt is evaluated, not in a separate encoder path. Both text branches share the same prompt learner and the same text encoder; what differs is the state of the context vectors and whether gradients flow. The zero-shot text prior $\mathbf{t}_0 = g(\mathbf{u}^{(0)})$ is the outcome of text encoder applied to the prompt at its initialization state $\mathbf{u}^{(0)}$, whose context vectors still hold the token embeddings of the template ``a photo of a $<y_c>$''; this feature is computed once before optimization and cached under a no-gradient context, so it never updates and carries no effective trainable parameters. It therefore equals the zero-shot CLIP text embedding of the template. The adapted branch $\mathbf{t}_a = g(\mathbf{u}^{(a)})$ applies the same encoder to the same prompt learner as its context vectors are optimized during training. 

Concretely, for vision modality the vanilla CLIP encoder $f(.)$ is weight-frozen and replicated to obtain frozen $f_0(.)$, and adapted $f_a(.)$ branches. Using these, we extract the zero-shot image prior
$\mathbf{z}_0 = f_{0}(\mathbf{x}) \in \mathbb{R}^d$ and the adapted image features $\mathbf{z}_a = f_{a}(\mathbf{x})$. We refer to the pair
$(\mathbf{z}_0, \mathbf{t}_0)$ as the zero-shot priors
and
$(\mathbf{z}_a, \mathbf{t}_a)$ as the adapted features. This makes the text zero-shot prior symmetric in role to the image zero-shot prior: just as $\mathbf{z}_0 = f_0(\mathbf{x})$ is the untouched CLIP image embedding, $\mathbf{t}_0 = g(\mathbf{u}^{(0)})$ is the untouched CLIP text embedding, both fixed throughout training. We note that all of these embeddings undergo $L_2$ normalization across the dimension $d$. The zero-shot priors preserve CLIP's robust pretrained geometry, while the adapted features capture task-specific semantics of the downstream dataset. Importantly, because $\mathbf{z}_0$ and $\mathbf{t}_0$ are captured with no-gradient evaluation and never recomputed, they receive no gradients and remain fixed throughout training, providing a stable semantic anchor for the fusion described next.

\subfour{Bilinear Similarity Fusion: }A key design choice in MM-Log is how to reconcile the stability of the zero-shot priors and the adaptability of the adapted features. Keeping the VLM entirely zero-shot maintains robust, general semantics but may fail to correct dataset-specific misalignment that harms OOD detection. Conversely, aggressive adapting can overfit to ID idiosyncrasies and degrade the pretrained geometry that makes VLMs broadly useful. We address this tension through bilinear similarity fusion: we compute cross-modal similarity using both the zero-shot features (capturing pretrained priors) and the adapted features (capturing ID-specific alignment), and then fuse these similarity signals into a single OOD score. This fusion operates directly at the logit level where detection decisions are made, making the method simple, modular, and compatible with standard VLM inference pipelines. Intuitively, zero-shot similarities provide a conservative semantic anchor, while adapted similarities sharpen ID agreement; their fusion yields a detector that is simultaneously well-calibrated and ID-sensitive.
\par With Bilinear similarity fusion, we form a fused representation for each modality by combining its zero-shot and adapted branches, $\mathbf{z} = \mathbf{z}_0 + \mathbf{z}_a$ and $\mathbf{t} = \mathbf{t}_0 + \mathbf{t}_a$. Computing the cross-modal similarity of these fused representations expands into four interaction terms:

$$
\begin{aligned}
S_\text{MM-Log}
&= \mathbf{z}^\top \mathbf{t} \\
&= (\mathbf{z}_{0} + \mathbf{z}_{a})^\top (\mathbf{t}_0 + \mathbf{t}_a) \\
&= \mathbf{z}_0^\top\mathbf{t}_0 + \mathbf{z}_0^\top\mathbf{t}_a
 + \mathbf{z}_a^\top\mathbf{t}_0 + \mathbf{z}_a^\top\mathbf{t}_a\\
&= S_{00} + S_{0a} + S_{a0} + S_{aa}\\
&= \sum_{i,j\in\{0,a\}} S_{ij}
\label{eq:lisf-decomp}
\end{aligned}$$

Consequently the MM-Log objective is a single cross-entropy applied to the fused logits:
\begin{equation}
\mathcal{L}_{\text{MM-Log}}
= \mathbb{E}_{(\mathbf{x}, y)\sim \mathbb{P}_{\text{ID}}}
\Bigg[
- \log
\frac{
e^{S_{\text{MM-Log}}^{\,y} / \tau
}}{
\sum_{i=1}^{C}
e^{S_{\text{MM-Log}}^{\,i} / \tau
}}
\Bigg]
\label{eq:loss}
\end{equation}
We deliberately supervise the fused score rather than each channel with its own cross-entropy: the softmax of a sum is not the product of per-channel softmaxes, and applying the loss at the fusion point is exactly what couples the channels. Gradients reach the trainable parameters only through the channels that contain an adapted component ($S_{aa}$, $S_{0a}$, $S_{a0}$), while the parameter-free $S_{00}$ enters the softmax as a fixed, input-dependent offset that shifts the per-class competition without itself receiving gradients. Following CLIP~\cite{radford2021learning}, all cosine similarities are scaled by the model's learned logit scale before the softmax; we absorb this scaling into $1/\tau$ in our notation.

\noindent\textbf{Why the cross-terms aid OOD detection?}
It is worth making explicit why the mixed terms, rather than $S_{aa}$ alone, are
what stabilizes detection under distribution shift. Because all embeddings are
$L_2$-normalized, each channel is a cosine agreement in $[-1,1]$ and the fused
score is their sum, $S_{\text{MM-Log}}\in[-4,4]$ (the learned logit scale
multiplies this by a positive constant and preserves all orderings).

Consider first an ID sample of class $c$. Training drives $\mathbf{z}_a$ and
$\mathbf{t}_a$ into tight alignment, so $S_{aa}$ is large; and because the loss
of Eq.~\eqref{eq:loss} is applied to the \emph{fused} score, gradients through
$S_{a0}$ pull the adapted image features toward the zero-shot text embeddings
while gradients through $S_{0a}$ pull the adapted prompts toward the zero-shot
image features. The zero-shot priors thus act as fixed targets that regularize
adaptation, so the cross-terms remain large as well and all four channels agree. Now consider an OOD sample. The adapted branch, optimized purely on ID data, can
still assign a spuriously high $S_{aa}$ to an input that exploits a shortcut or
dataset-specific artifact -- precisely the overconfidence failure mode of
adapted-only detectors. The zero-shot priors, however, were never adapted to that
shortcut, so the cross-terms $S_{0a}=\mathbf{z}_0^\top\mathbf{t}_a$ and
$S_{a0}=\mathbf{z}_a^\top\mathbf{t}_0$ compare an adapted embedding in one
modality against a \emph{non-adapted} embedding in the other. An input that fools
the adapted branch must therefore additionally agree with the zero-shot CLIP
geometry to keep the summed score high. The full MM-Log procedure is provided in Algorithm.~\ref{alg:mmlog}.

\subfour{Theoretical Justification: }
By bilinearity, the four separately computed channels sum to a single inner
product,
\begin{equation}
S_{\text{MM-Log}}
=\underbrace{\big(\mathbf{z}_0+\mathbf{z}_a\big)^{\!\top}}_{\mathbf{Z}\ \text{(analysis only)}}
 \underbrace{\big(\mathbf{t}_0+\mathbf{t}_a\big)}_{\mathbf{T}\ \text{(analysis only)}},
\label{eq:factor}
\end{equation}
where $\mathbf{Z},\mathbf{T}$ are used only for analysis; the model never forms
them. Since their summands are unit vectors, their lengths measure exactly how
far adaptation has moved from the pretrained priors:
$\lVert\mathbf{Z}\rVert=2\cos\frac{\alpha_z}{2}$ and
$\lVert\mathbf{T}\rVert=2\cos\frac{\alpha_t}{2}$, where
$\alpha_z=\angle(\mathbf{z}_0,\mathbf{z}_a)$ and
$\alpha_t=\angle(\mathbf{t}_0,\mathbf{t}_a)$. 

\begin{proposition}
\label{prop:margin}
Let $\delta=\max(\alpha_z,\alpha_t)$ denote the larger zero-shot--adapted drift.
Then
\begin{equation}
S_{\text{\emph{MM-Log}}}\;\le\;4\cos\tfrac{\alpha_z}{2}\cos\tfrac{\alpha_t}{2}
\;\le\;4\cos\tfrac{\delta}{2}\;=\;4-\tfrac12\delta^{2}+O(\delta^{4}).
\label{eq:margin}
\end{equation}
If, in addition, every cross-modal pair satisfies
$\angle(\mathbf{z}_i,\mathbf{t}_j)\le\epsilon$ for all $i,j\in\{0,a\}$, then
$S_{\text{\emph{MM-Log}}}\ge 4\cos\epsilon=4-2\epsilon^{2}+O(\epsilon^{4})$.
\end{proposition}

\begin{proof}
\emph{Upper bound.} Assume without loss of generality $\alpha_z\ge\alpha_t$, so
$\delta=\alpha_z$. Then
\begin{align}
S_{\text{MM-Log}}
&=\mathbf{Z}^\top\mathbf{T}
&&\text{by \eqref{eq:factor}} \nonumber\\
&\le\lVert\mathbf{Z}\rVert\,\lVert\mathbf{T}\rVert
&&\text{(Cauchy--Schwarz)} \nonumber\\
&=4\cos\tfrac{\alpha_z}{2}\cos\tfrac{\alpha_t}{2}
&&\text{(half-angle identity)} \nonumber\\
&\le 4\cos\tfrac{\delta}{2}
&&\text{since }\cos\tfrac{\alpha_t}{2}\le 1 \nonumber\\
&=4-\tfrac12\delta^{2}+O(\delta^{4})
&&\text{(Taylor expansion).} \nonumber
\end{align}

\emph{Lower bound.} Each channel equals $\cos\angle(\mathbf{z}_i,\mathbf{t}_j)$.
If every cross-modal pair satisfies $\angle(\mathbf{z}_i,\mathbf{t}_j)\le\epsilon$,
monotonicity of $\cos$ on $[0,\pi]$ gives $S_{ij}\ge\cos\epsilon$ for each $i,j$,
and summing the four channels yields
$S_{\text{MM-Log}}\ge4\cos\epsilon=4-2\epsilon^{2}+O(\epsilon^{4})$.
\end{proof}

Proposition~\ref{prop:margin} formalizes the argument above. An input whose
adaptation has drifted by $\delta$ in either modality is capped at
$4-\frac12\delta^2$ and cannot recover a high fused score however confidently the
adapted branch aligns it, since that confidence enters only through $S_{aa}$;
an input whose four features mutually agree to within $\epsilon$ retains at least
$4-2\epsilon^2$. Whenever $\delta>2\epsilon$ the first bound falls below the
second, guaranteeing a separating threshold. The cross-terms are what make this
possible: $S_{aa}$ involves neither $\mathbf{z}_0$ nor $\mathbf{t}_0$, so it is
invariant to the zero-shot geometry and cannot distinguish two inputs that differ
only in how far they have drifted from it. The bound establishes \emph{sensitivity} rather than detection: whether OOD
inputs actually exhibit larger $\delta$ than ID inputs is an empirical property
of the data and the trained model. Tables~\ref{tab:4} and~\ref{tab:5} confirm the
predicted ordering, with variants omitting the cross-terms (e.g.,
$S_{00}+S_{aa}$) most vulnerable and the progressive addition of $S_{0a}$,
$S_{a0}$, and $S_{00}$ yielding the most reliable separation.

\subfour{Inference: }
During inference, we score a test input with the free energy of the fused logits~\cite{liu2020energy}:
\begin{equation}
s(\mathbf{x})
=
T \cdot \log \sum_{c=1}^{C} e^{\,S^{\,c}_{\text{MM-Log}} / T},
\label{eq:energy}
\end{equation}
with energy temperature $T$, and plug $s(\mathbf{x})$ into the detector of Eq.~\eqref{eq:2}. Energy aggregates evidence over all classes rather than committing to the single best match, which suits the fused logit space: an ID input accumulates consistently high agreement across the four channels for its correct class, whereas an OOD input that partially fools one channel rarely sustains high fused logits across the class set, so its logsumexp remains low.

\begin{algorithm}[ht!]
\caption{MM-Log: finetuning and OOD inference}
\label{alg:mmlog}
\begin{algorithmic}[1]
\Require pretrained CLIP encoders $f(\cdot),g(\cdot)$; ID data
$(\mathbf{x},y)\sim\mathbb{P}_{\text{ID}}$ over classes $\{1,\dots,C\}$;
softmax temperature $\tau$; energy temperature $T$; threshold $\gamma$
\Statex \textbf{\emph{Zero-shot priors (frozen, computed once)}}
\State $f_0 \gets$ frozen copy of $f$;\quad
       $\mathbf{u}^{(0)} \gets$ initialization prompt (frozen)
\State $\mathbf{t}_0^{\,c} \gets g(\mathbf{u}^{(0)})$ for $c=1,\dots,C$
  \Comment{cache zero-shot text features; all features $L_2$-normalized}
\Statex \textbf{\emph{Training on ID data}}
\State initialize adapted encoder $f_a$ and adapted prompt $\mathbf{u}^{(a)}$
\Repeat
  \For{minibatch $(\mathbf{x},y)\sim\mathbb{P}_{\text{ID}}$}
    \State $\mathbf{z}_0 \gets f_0(\mathbf{x})$,\quad $\mathbf{z}_a \gets f_a(\mathbf{x})$
    \State $\mathbf{t}_a^{\,c} \gets g(\mathbf{u}^{(a)})$ for $c=1,\dots,C$
    \State $S_{\text{MM-Log}}^{\,c} \gets
           \sum_{i,j\in\{0,a\}} \mathbf{z}_i^\top \mathbf{t}_j^{\,c}$,
           \quad $c=1,\dots,C$
    \State $\mathcal{L}_{\text{MM-Log}} \gets
           -\log \dfrac{e^{S_{\text{MM-Log}}^{\,y}/\tau}}
                       {\sum_{i=1}^{C} e^{S_{\text{MM-Log}}^{\,i}/\tau}}$
    \State update $f_a,\mathbf{u}^{(a)}$ via $\nabla\mathcal{L}_{\text{MM-Log}}$
           \Comment{$f_0,\mathbf{u}^{(0)}$ frozen}
  \EndFor
\Until{converged}
\Statex \textbf{\emph{OOD detection for a test input $\mathbf{x}$}}
\State compute $S_{\text{MM-Log}}^{\,c}$ for $c=1,\dots,C$ as in line 8
\State $s(\mathbf{x}) \gets T\,\log\!\sum_{c=1}^{C}
       \exp\!\big(S_{\text{MM-Log}}^{\,c}/T\big)$
\State \Return $G(\mathbf{x}) = \text{ID if } s(\mathbf{x})\ge\gamma \text{ else OOD}$
\end{algorithmic}
\end{algorithm}

\section{Experimental setup}
Following the existing studies in VLM based OOD detection \cite{miao2025auxiliary, ming2022delving, lu2025fa}, we utilize CLIP ViT-B/16 as the backbone model and adapt it for 50 epochs on a downstream image dataset. The adapted prompt uses $N{=}4$ context vectors initialized from the token embeddings of the template ``a photo of a $y_c$'', following the CoOp initialization scheme~\cite{zhou2022learning}. The hyperparameters used for training are as follows: batch size is 160, learning rate is $2e-3$, SGD optimizer (momentum 0.9, and weight decay $5e-4$) with a cosine scheduler. Cosine similarities are scaled by CLIP's learned logit scale, and the softmax and energy temperature are set to $\tau = 1$, and $T{=}1$ respectively. All the experiments including training and testing were conducted on a NVIDIA A100 GPU cluster. For statistical significance of the experiment, we train the models for 3 trials with different random seeds and report the averaged test metrics over these trials. We provide comprehensive experimental setup including details about algorithm, datasets, baselines in the Supplementary material.

\subfour{Datasets: }
We train MM-Log on the ImageNet-1k recognition benchmark as the ID dataset, and evaluate OOD detection across three regimes of increasing difficulty. \textbf{Far-OOD} uses iNaturalist, SUN, Places, and Textures, which are semantically and visually distant from ImageNet classes (fine-grained natural species, scene imagery, and surface textures). \textbf{Near-OOD} uses OpenImage-O, NINCO, and ImageNet-O, whose classes are visually or semantically close to the ID categories while sharing no label overlap, making them substantially harder to reject. \textbf{Hard-OOD} pairs the ImageNet-10, ImageNet-20, and ImageNet-100 subsets against one another (e.g., ImageNet-10 vs. ImageNet-20), so that ID and OOD distributions overlap in semantic content and the detector must separate closely related concepts.
\subfour{Baselines and Metrics: }We compare MM-Log's performance against popular baselines, some of them being state-of-the-art in OOD detection. These include; Coop \cite{zhou2022learning}, LoCoop \cite{miyai2023locoop}, CLIPN \cite{wang2023clipn}, IDLike \cite{bai2024id}, NegLabel \cite{jiang2024negative}, LAPT \cite{zhang2024lapt}, SCT \cite{yu2024self}, APT \cite{miao2025auxiliary}, CMA \cite{kim2025enhanced}, and FA \cite{lu2025fa}. To evaluate OOD detection performance, we use Area Under the Receiving Operator Characteristics Curve (AUROC), and False Positive Rate @ 95\% of True Positive Rate (FPR95). Likewise, we also use ID accuracy to compare the classification performance of the model.

\section{Results}
\subfour{Performance on Far-OOD dataset:}
Table~\ref{tab:1} reports Far-OOD results with ImageNet-1k as ID. MM-Log achieves the best average AUROC and FPR95; although it is not top on iNaturalist, it outperforms all baselines on SUN, Places, and Textures. The second-best method, LAPT~\cite{zhang2024lapt}, relies on label-driven retrieval of textual descriptions from web-scale data, and its effectiveness hinges on the coverage of the retrieved labels; MM-Log requires no such external retrieval. Against the most design-similar baseline, FA~\cite{lu2025fa}, which does not model zero-shot priors in the image branch, MM-Log improves average AUROC by more than $2\%$ and nearly halves FPR95.

\begin{table*}[ht!]
    \caption{Comparison of OOD Detection metrics trained on ImageNet-1k benchmark as ID and tested on Far-OOD datasets. Results of methods represented with $\dagger$ are directly reported form their corresponding paper.}
    \label{tab:1}
    \centering
    \resizebox{\textwidth}{!}{%
    \begin{tabular}{lccccccccccc}
        \toprule
        
        \multirow{3}{*}{\bf Method} 
        & \multirow{3}{*}{\bf Venues} & \multicolumn{10}{c}{\bf OOD Dataset} \\
        \cmidrule{3-12}
        & & \multicolumn{2}{c}{\bf iNaturalist}  
        & \multicolumn{2}{c}{\bf SUN} 
        & \multicolumn{2}{c}{\bf Places}
        & \multicolumn{2}{c}{\bf Textures} 
        & \multicolumn{2}{c}{\bf Average} \\

        & &AUROC $\uparrow$ &FPR95 $\downarrow$ 
        & AUROC $\uparrow$ & FPR95 $\downarrow$ 
        & AUROC $\uparrow$ & FPR95 $\downarrow$ 
        & AUROC $\uparrow$ & FPR95 $\downarrow$ 
        & AUROC $\uparrow$ & FPR95 $\downarrow$ \\
        
        \midrule
        
        Coop &IJCV'22& $91.47\scriptstyle{\pm 2.12}$ & $41.14\scriptstyle{\pm 9.39}$ 
        & $91.68\scriptstyle{\pm0.83}$ & $ 39.06\scriptstyle{\pm 4.17}$ 
        & $89.16\scriptstyle{\pm 1.28}$ & $45.38\scriptstyle{\pm 4.88}$ 
        & $87.82\scriptstyle{\pm1.12 }$ & $51.37\scriptstyle{\pm 3.09}$ 
        & $90.03\scriptstyle{\pm 0.32}$ & $44.24\scriptstyle{\pm 1.39}$ \\
        
        LoCoop &NeurIPS'23& $92.78\scriptstyle{\pm 0.93}$ & $36.64\scriptstyle{\pm 4.87}$ 
        & $93.50\scriptstyle{\pm0.92}$ & $ 31.86\scriptstyle{\pm 3.79}$ 
        & $90.70\scriptstyle{\pm 0.97}$ & $38.81\scriptstyle{\pm 3.37}$ 
        & $89.43\scriptstyle{\pm0.68 }$ & $48.11\scriptstyle{\pm 2.31}$ 
        & $91.59\scriptstyle{\pm 0.58}$ & $38.85\scriptstyle{\pm 2.67}$ \\
        
        CLIPN$\dagger$ &ICCV'23& $95.27\scriptstyle{\pm0.00 }$ & $23.94\scriptstyle{\pm 0.00 }$ 
        & $93.93\scriptstyle{\pm 0.00}$ & $26.17\scriptstyle{\pm 0.00}$ 
        & $90.93\scriptstyle{\pm 0.00}$ & $40.83\scriptstyle{\pm 0.00}$ 
        & $92.28\scriptstyle{\pm 0.00}$ & $33.45\scriptstyle{\pm 0.00}$ 
        & $93.10\scriptstyle{\pm 0.00}$ & $31.10\scriptstyle{\pm 0.00}$ \\

        IDLike &CVPR'24& $96.68\scriptstyle{\pm0.31 }$ & $17.73\scriptstyle{\pm 1.91}$ 
        & $89.53\scriptstyle{\pm 0.58}$ & $48.17\scriptstyle{\pm 1.39}$ 
        & $88.27\scriptstyle{\pm 2.27}$ & $50.43\scriptstyle{\pm6.58 }$ 
        & $93.25\scriptstyle{\pm2.16 }$ & $29.12\scriptstyle{\pm 7.64}$ 
        & $91.93\scriptstyle{\pm 1.18}$ & $36.36\scriptstyle{\pm 3.86}$ \\
        
        NegLabel $\dagger$ &ICLR'24& $\underline{99.49\scriptstyle{\pm 0.00}}$ & $\underline{1.91\scriptstyle{\pm 0.00}}$ 
        & $95.49\scriptstyle{\pm 0.00}$ & $20.53\scriptstyle{\pm 0.00}$ 
        & $91.64\scriptstyle{\pm 0.00}$ & $35.59\scriptstyle{\pm 0.00}$ 
        & $90.22\scriptstyle{\pm 0.00}$ & $43.56\scriptstyle{\pm 0.00}$ 
        & $94.21\scriptstyle{\pm 0.00}$ & $25.40\scriptstyle{\pm 0.00}$ \\
        
        LAPT$\dagger$ &ECCV'24& $\bf{99.63\scriptstyle{\pm 0.00}}$ & $\mathbf{1.16\scriptstyle{\pm 0.00}}$ 
        & $\underline{96.01\scriptstyle{\pm 0.00}}$ & $\underline{19.12\scriptstyle{\pm 0.00}}$ 
        & $92.01\scriptstyle{\pm 0.00}$ & $33.01\scriptstyle{\pm 0.00}$ 
        & $91.06\scriptstyle{\pm 0.00}$ & $40.32\scriptstyle{\pm 0.00}$ 
        & $\underline{94.68\scriptstyle{\pm 0.00}}$ & $\underline{23.40\scriptstyle{\pm 0.00}}$ \\
        
        SCT &NeurIPS'24&  $ 91.77\scriptstyle{\pm2.40 }$ &  $ 41.93\scriptstyle{\pm12.17 }$ 
        &  $93.76 \scriptstyle{\pm 0.39}$ &  $ 30.39\scriptstyle{\pm2.17 }$ 
        &  $90.78 \scriptstyle{\pm0.34 }$ &  $ 38.73\scriptstyle{\pm2.54 }$ 
        &  $ 88.96\scriptstyle{\pm 1.16}$ &  $ 46.78\scriptstyle{\pm 3.51}$ &  $ 91.32\scriptstyle{\pm 0.90}$ &  $ 39.46\scriptstyle{\pm 3.39}$ \\
        
        APT$\dagger$ &ICCV'25& $97.79\scriptstyle{\pm0.00}$ & $9.70\scriptstyle{\pm0.00}$ 
        & $95.52\scriptstyle{\pm0.00}$ & $20.12\scriptstyle{\pm0.00}$ 
        & $\underline{92.84\scriptstyle{\pm0.00}}$ & $\underline{28.54\scriptstyle{\pm0.00}}$ 
        & $88.43\scriptstyle{\pm0.00}$ & $45.78\scriptstyle{\pm0.00}$ 
        & $93.64\scriptstyle{\pm0.00}$ & $26.03\scriptstyle{\pm0.00}$ \\
        
        CMA$\dagger$ &CVPR'25& $ 95.67\scriptstyle{\pm0.00}$ & $ 22.81\scriptstyle{\pm0.00}$ 
        & $ 91.64\scriptstyle{\pm0.00}$ & $ 40.60\scriptstyle{\pm0.00}$ 
        & $ 88.41\scriptstyle{\pm0.00}$ & $ 48.71\scriptstyle{\pm0.00}$ 
        & $89.68 \scriptstyle{\pm0.00}$ & $45.23 \scriptstyle{\pm0.00}$ 
        & $91.35 \scriptstyle{\pm0.00}$ & $ 39.33\scriptstyle{\pm0.00}$ \\
        
        FA &ICCV'25& $94.72 \scriptstyle{\pm 0.27}$ & $25.50 \scriptstyle{\pm 2.72}$ 
        & $92.40 \scriptstyle{\pm 0.95}$ & $36.24 \scriptstyle{\pm 3.15}$ 
        & $91.99 \scriptstyle{\pm 0.68}$ & $35.38 \scriptstyle{\pm 2.75}$ 
        & $\underline{93.95 \scriptstyle{\pm 0.19}}$ & $\underline{28.34 \scriptstyle{\pm 0.87}}$ 
        & $93.25 \scriptstyle{\pm 0.32}$ & $31.37 \scriptstyle{\pm 1.07}$ \\

        \midrule
        \rowcolor{blue!20}
        MM-Log &-& $97.49 \scriptstyle{\pm 0.39}$ & $10.53 \scriptstyle{\pm 2.09}$ 
        & $\mathbf{96.13 \scriptstyle{\pm 0.59}}$ & $\mathbf{15.12 \scriptstyle{\pm 3.85}}$ 
        & $\mathbf{94.60 \scriptstyle{\pm 0.75}}$ & $\mathbf{20.89 \scriptstyle{\pm 3.79}}$ 
        & $\mathbf{94.68 \scriptstyle{\pm 1.59}}$ & $\mathbf{27.61 \scriptstyle{\pm 6.29}}$ 
        & $\mathbf{95.72\scriptstyle{\pm 0.83}}$ & $\mathbf{18.53\scriptstyle{\pm 4.00}}$ \\
        \bottomrule
    \end{tabular}
    }
\end{table*}

\begin{table*}[ht!]
    \caption{Comparison of OOD Detection metrics trained on ImageNet-1k benchmark as ID and tested on Near-OOD datasets. The best and second best results are reported in bold and underline respectively.}
    \centering
    \label{tab:2}
    \resizebox{\textwidth}{!}{%
    \begin{tabular}{lccccccccc}
        \toprule
        \multirow{3}{*}{\bf Method} 
        & \multirow{3}{*}{\bf Venues} & \multicolumn{8}{c}{\bf OOD Dataset} \\
        \cmidrule{3-10}
        & & \multicolumn{2}{c}{\bf OpenImage-O}  
        & \multicolumn{2}{c}{\bf NINCO} 
        & \multicolumn{2}{c}{\bf ImageNet-O}
        & \multicolumn{2}{c}{\bf Average} \\
        & &AUROC $\uparrow$ &FPR95 $\downarrow$
        & AUROC $\uparrow$ & FPR95 $\downarrow$ 
        & AUROC $\uparrow$ & FPR95 $\downarrow$ 
        & AUROC $\uparrow$ & FPR95 $\downarrow$ \\
        
        \midrule

        COOP &IJCV'22& $91.99\scriptstyle{\pm 0.22}$ & $37.95\scriptstyle{\pm 0.59}$ 
        & $73.00\scriptstyle{\pm 0.59}$ & $78.00\scriptstyle{\pm 0.23}$ 
        & $81.34\scriptstyle{\pm 0.83}$ & $70.14\scriptstyle{\pm 0.86}$ 
        & $82.11\scriptstyle{\pm 0.48}$ & $62.03\scriptstyle{\pm 0.38}$ \\

        LoCoop &NeurIPS'23& $92.22\scriptstyle{\pm 0.25}$ & $32.23\scriptstyle{\pm 1.00}$ 
        & $75.35\scriptstyle{\pm 0.88}$ & $72.04\scriptstyle{\pm 0.81}$ 
        & $78.44\scriptstyle{\pm 0.94}$ & $67.35\scriptstyle{\pm 1.19}$ 
        & $81.99\scriptstyle{\pm 0.58}$ & $57.21\scriptstyle{\pm 0.96}$ \\

        IDLike &CVPR'24& $87.89\scriptstyle{\pm 0.45}$ & $54.43\scriptstyle{\pm 2.35}$ 
        & $69.32\scriptstyle{\pm 0.4}$ & $78.93\scriptstyle{\pm 1.79}$ 
        & $67.45\scriptstyle{\pm 0.86}$ & $84.08\scriptstyle{\pm 1.45}$ 
        & $74.89\scriptstyle{\pm 0.28}$ & $72.48\scriptstyle{\pm 0.89}$ \\

        SCT &NeurIPS'24& $92.04\scriptstyle{\pm 0.33}$ & $37.28\scriptstyle{\pm 1.16}$ 
        & $71.09\scriptstyle{\pm 1.19}$ & $78.51\scriptstyle{\pm 0.86}$ 
        & $81.29\scriptstyle{\pm 0.15}$ & $71.45\scriptstyle{\pm 0.54}$ 
        & $81.48\scriptstyle{\pm 0.52}$ & $62.41\scriptstyle{\pm 0.73}$ \\
        
        CMA &CVPR'25& $\underline{94.32\scriptstyle{\pm0.00 }}$ & $\underline{28.85\scriptstyle{\pm0.00 }}$ 
        & $\underline{82.18\scriptstyle{\pm0.00 }}$ & $\underline{65.64\scriptstyle{\pm 0.00}}$ 
        & $\underline{84.33\scriptstyle{\pm0.00 }}$ & $\underline{63.55\scriptstyle{\pm0.00 }}$ 
        & $\underline{86.94\scriptstyle{\pm 0.00}}$ & $\underline{52.68\scriptstyle{\pm0.00 }}$ \\

        FA &ICCV'25& $91.84 \scriptstyle{\pm 0.45}$ & $37.93 \scriptstyle{\pm 1.65}$ 
        & $77.77 \scriptstyle{\pm 0.94}$ & $70.44 \scriptstyle{\pm 1.24}$ 
        & $82.71 \scriptstyle{\pm 0.30}$ & $63.84 \scriptstyle{\pm 1.36}$ 
        & $84.08 \scriptstyle{\pm 0.47}$ & $57.40 \scriptstyle{\pm 1.34}$ \\

        \midrule
        \rowcolor{blue!20}
        MM-Log &-& $\mathbf{96.07 \scriptstyle{\pm 0.68}}$ &$\mathbf{17.66 \scriptstyle{\pm 3.33}}$ 
        & $\mathbf{83.39 \scriptstyle{\pm 0.50}}$ & $\mathbf{59.38 \scriptstyle{\pm 2.19}}$ 
        & $\mathbf{87.77 \scriptstyle{\pm 2.30}}$ & $\mathbf{46.93 \scriptstyle {\pm 6.24}}$ 
        & $\mathbf{89.27\scriptstyle{\pm1.16 }}$ & $\mathbf{41.32\scriptstyle{\pm3.92 }}$ \\
        \bottomrule
    \end{tabular}
    }
\end{table*}

\begin{table*}[ht!]
    \caption{Hard OOD on test for ImageNet-$n$ variants. The best and second best results are reported in bold and underline respectively.}
    \centering
    \label{tab:3}
    \resizebox{\textwidth}{!}{%
    \begin{tabular}{lccccccccccc}
        \toprule
        \multirow{3}{*}{\bf Method} & \bf ID & \multicolumn{2}{c}{\bf ImageNet-10}  
        & \multicolumn{2}{c}{\bf ImageNet-20} 
        & \multicolumn{2}{c}{\bf ImageNet-10}
        & \multicolumn{2}{c}{\bf ImageNet-100} & \multicolumn{2}{c}{\bf Average} \\
        
        & \bf OOD & \multicolumn{2}{c}{\bf ImageNet-20}  
        & \multicolumn{2}{c}{\bf ImageNet-10} 
        & \multicolumn{2}{c}{\bf ImageNet-100}
        & \multicolumn{2}{c}{\bf ImageNet-10} \\
        & &AUROC $\uparrow$ &FPR95 $\downarrow$
        & AUROC $\uparrow$ & FPR95 $\downarrow$ 
        & AUROC $\uparrow$ & FPR95 $\downarrow$ 
        & AUROC $\uparrow$ & FPR95 $\downarrow$
         & AUROC $\uparrow$ & FPR95 $\downarrow$\\
        
        \midrule
        LoCoop &-&92.75 &28.20 & 92.34 & 34.40 & 93.00 & 30.08 & 81.97 & 61.40 & 90.01 & 38.52\\
        
        SCT &-& 94.33& 25.10 & 94.95 & 25.00 & 93.90 &26.64 & 82.60 & 57.80  &91.44& 33.63\\
        
        NegLabel &-& \underline{98.86}  & \underline{5.10} &\bf98.81& \bf4.60 &\bf99.51 &\bf1.68& 90.19 &40.20 &96.84 &\underline{12.89}   \\

        IDLike &-& 97.32 & 9.43 & 97.96 & 10.28 &96.43 &9.81 & 88.03& 45.67 &94.93&18.79\\
        
        FA &-& 98.36 & 4.36 & 98.13& 12.05& 97.81& 6.84 & \underline{94.38}& \underline{33.76} &\underline{97.17}&14.25 \\
        
        \midrule
        \rowcolor{blue!20}
        MM-Log &-& \bf 99.38 & \bf 0.87
        & \underline{98.17} & \underline{11.79}
        & \underline{99.06}& \underline{5.46} 
        & \bf 95.37 & \bf 22.55 &\bf97.99& \bf10.16\\
        \bottomrule
    \end{tabular}
    }
\end{table*}
\subfour{Performance on Near-OOD dataset:}
In Table~\ref{tab:2}, MM-Log achieves the best AUROC and FPR95 across all Near-OOD datasets, improving over the strongest baseline, CMA, by a substantial average margin. We attribute these gains to MM-Log’s expanded similarity formulation, which jointly models frozen and learnable components of both image and text representations.

\begin{table*}[ht!]
    \caption{Comparison of OOD detection results across different fusion types for Near-OOD datasets. Subscripts denote the two branches contributing to each similarity term ($0$: zero-shot, $a$: adapted; e.g., $S_{0a}=\mathbf{z}_0^\top\mathbf{t}_a$). Rows highlighted in color represent performance without logit fusion. Best and second-best results are in bold and underline respectively.}
    \centering
    \label{tab:4}
    \resizebox{\textwidth}{!}{%
    \begin{tabular}{lcccccccccc}
        \toprule
        \multirow{3}{*}{\bf Fusion Type} 
        & \multicolumn{10}{c}{\bf OOD Dataset} \\
        \cmidrule{3-11}
        & \multicolumn{2}{c}{\bf iNaturalist}  
        & \multicolumn{2}{c}{\bf SUN} 
        & \multicolumn{2}{c}{\bf Places}
        & \multicolumn{2}{c}{\bf Textures} 
        & \multicolumn{2}{c}{\bf Average} \\

        & AUROC $\uparrow$ &FPR95 $\downarrow$ 
        & AUROC $\uparrow$ & FPR95 $\downarrow$ 
        & AUROC $\uparrow$ & FPR95 $\downarrow$ 
        & AUROC $\uparrow$ & FPR95 $\downarrow$ 
        & AUROC $\uparrow$ & FPR95 $\downarrow$ \\
        
        \midrule
        \rowcolor{blue!20}
        $S_{00}$ & $94.16\scriptstyle{\pm 1.12}$ & $31.95\scriptstyle{\pm 0.45}$ & $92.55\scriptstyle{\pm 3.44}$ & $37.22\scriptstyle{\pm 2.34}$& $90.10\scriptstyle{\pm 3.12}$ & $42.98\scriptstyle{\pm 1.18}$ & $85.83\scriptstyle{\pm 2.73}$ & $58.35\scriptstyle{\pm 0.98}$ & $90.66\scriptstyle{\pm 2.60}$ & $42.63\scriptstyle{\pm1.23 }$ \\
        \rowcolor{blue!20}
        $S_{aa}$ & $91.35\scriptstyle{\pm 2.33}$ &$41.68\scriptstyle{\pm 1.27}$& $90.79\scriptstyle{\pm 2.70}$ & $43.79\scriptstyle{\pm 3.42}$ & $87.90\scriptstyle{\pm 1.83}$ & $50.47\scriptstyle{\pm 0.87}$& $88.76\scriptstyle{\pm 2.12}$ & $47.59\scriptstyle{\pm 3.40}$ & $89.70\scriptstyle{\pm 2.24}$ & $45.88\scriptstyle{\pm 2.23}$ \\

        \midrule

        $S_{00}+S_{aa}$ &$93.40\scriptstyle{\pm 3.47}$ & $34.39\scriptstyle{\pm 2.83}$ & $91.50\scriptstyle{\pm 1.77}$ & $43.36\scriptstyle{\pm 2.16}$ & $88.86\scriptstyle{\pm 0.93}$ & $49.97\scriptstyle{\pm 1.06}$ & $86.23\scriptstyle{\pm 3.81}$ & $58.71\scriptstyle{\pm 2.77}$ & $90.00\scriptstyle{\pm 2.49}$ & $46.61\scriptstyle{\pm 2.20}$ \\
        
        $S_{0a}+S_{aa}$ & $95.75\scriptstyle{\pm 0.45}$ & $17.79\scriptstyle{\pm1.47}$ 
        & $95.85\scriptstyle{\pm 1.89}$ & $16.38\scriptstyle{\pm3.60}$
        & $94.10\scriptstyle{\pm 0.79}$ & $23.10\scriptstyle{\pm3.54}$
        & $92.72\scriptstyle{\pm 0.82}$ & $28.23\scriptstyle{\pm3.22}$ 
        & $94.60\scriptstyle{\pm0.98 }$ & $21.37\scriptstyle{\pm2.95}$ \\
        
        $S_{a0}+S_{aa}$ & $94.82\scriptstyle{\pm 3.67}$ & $25.82\scriptstyle{\pm4.48}$ 
        & $90.39\scriptstyle{\pm 3.73}$ & $47.01\scriptstyle{\pm4.82}$
        & $89.32\scriptstyle{\pm 2.79}$ & $46.88\scriptstyle{\pm4.76}$
        & $87.79\scriptstyle{\pm 2.86}$ & $35.67\scriptstyle{\pm4.73}$ 
        & $90.85\scriptstyle{\pm 6.16}$ & $41.77\scriptstyle{\pm4.23}$ \\

        $S_{0a}+S_{aa}+S_{a0}$ & $\underline{96.39\scriptstyle{\pm 0.77}}$ & $\underline{14.06\scriptstyle{\pm3.98}}$ 
        & $\underline{96.01\scriptstyle{\pm 0.77}}$ & $\underline{16.01\scriptstyle{\pm3.30}}$
        & $\underline{94.18\scriptstyle{\pm 0.98}}$ & $\underline{21.29\scriptstyle{\pm3.32}}$
        & $\underline{93.66\scriptstyle{\pm 1.24}}$ & $\underline{28.10\scriptstyle{\pm3.24}}$ 
        & $\underline{95.06\scriptstyle{\pm 0.94}}$ & $\underline{19.86\scriptstyle{\pm}3.46}$ \\

        $S_{0a}+S_{aa}+S_{a0}+S_{00}$ & $\mathbf{97.49 \scriptstyle{\pm 0.39}}$ & $\mathbf{10.53 \scriptstyle{\pm 2.09}}$ 
        & $\mathbf{96.13 \scriptstyle{\pm 0.59}}$ & $\mathbf{15.12 \scriptstyle{\pm 3.85}}$ 
        & $\mathbf{94.60 \scriptstyle{\pm 0.75}}$ & $\mathbf{20.89 \scriptstyle{\pm 3.79}}$ 
        & $\mathbf{94.68 \scriptstyle{\pm 1.59}}$ & $\mathbf{27.61 \scriptstyle{\pm 6.29}}$ 
        & $\mathbf{95.72\scriptstyle{\pm 0.83}}$ & $\mathbf{18.53\scriptstyle{\pm 4.00}}$ \\
        
        \bottomrule
    \end{tabular}
    }
\end{table*}

\subfour{Performance on Hard-OOD dataset:}
Table~\ref{tab:3} presents results on the challenging ImageNet-$n$ variants. MM-Log achieves the best average AUROC and FPR95. NegLabel occasionally surpasses it on individual splits, but by small margins and only through explicitly defined negative labels whose quality may not transfer across datasets; MM-Log needs no such supervision and remains reliable even under heavy semantic overlap.

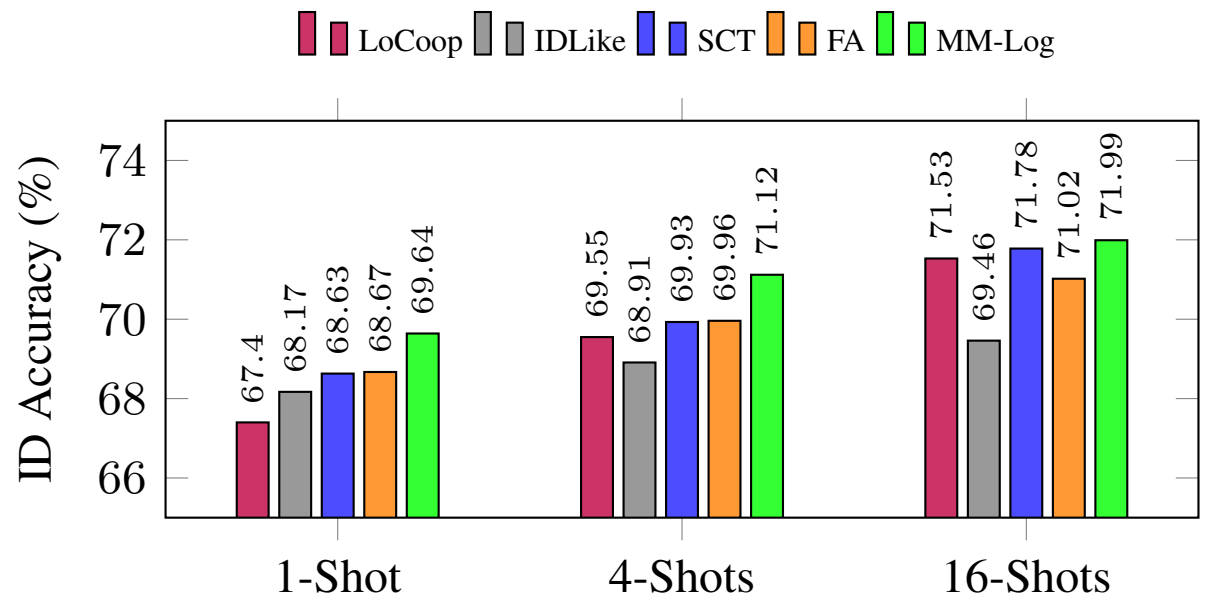
\begin{figure}[ht!]
\centering
\begin{tikzpicture}
\begin{axis}[
    width=\linewidth,
    height=0.5\linewidth,
    ybar,
    bar width=6pt,
    ymin=65, ymax=75,
    ylabel={ID Accuracy (\%)},
    symbolic x coords={1-Shot, 4-Shots, 16-Shots},
    xtick=data,
    xticklabel style={font=\small},
    ylabel style={font=\small},
    tick label style={font=\small},
    nodes near coords,
    nodes near coords align={vertical},
    nodes near coords style={
        font=\tiny,
        rotate=90,
        anchor=west
    },
    legend style={
        at={(0.5,1.1)},
        anchor=south,
        legend columns=5,
        font=\tiny,
        draw=none
    },
    enlarge x limits=0.25,
]

\addplot[fill=purple!80] coordinates {
    (1-Shot,67.40) (4-Shots,69.55) (16-Shots,71.53)
};

\addplot[fill=gray!80] coordinates {
    (1-Shot,68.17) (4-Shots,68.91) (16-Shots,69.46)
};

\addplot[fill=blue!70] coordinates {
    (1-Shot,68.63) (4-Shots,69.93) (16-Shots,71.78)
};

\addplot[fill=orange!80] coordinates {
    (1-Shot,68.67) (4-Shots,69.96) (16-Shots,71.02)
};

\addplot[fill=green!80] coordinates {
    (1-Shot,69.64) (4-Shots,71.12) (16-Shots,71.99)
};

\legend{LoCoop, IDLike, SCT, FA, MM-Log}

\end{axis}
\end{tikzpicture}
\caption{Comparison of ID classification accuracy of MM-Log with different baselines trained on ImageNet-1k dataset under different few-shot settings.}
\label{fig:id_acc}
\end{figure}

\subfour{Impact on ID-Accuracy:}
Figure~\ref{fig:id_acc} compares ID classification accuracy of MM-Log on ImageNet-1k under different few-shot settings (1-shot, 4-shots, and 16-shots) across five methods. Here, we can see that MM-Log consistently achieves the highest accuracy in all shot regimes, with particularly notable improvements in the low-shot setting. These gains can be attributed to MM-Log’s explicit modeling of fine-grained cross-modal interactions between zero-shot and adapted representations. By decomposing both image and text embeddings into zero-shot and adapted components and jointly optimizing all four interaction terms, MM-Log preserves the strong semantic priors of the pretrained model while enabling effective task-specific adaptation with limited labeled data. This expanded similarity formulation leads to improved alignment for ID image–text pairs without overfitting, resulting in better sample efficiency and more stable performance as the number of training examples increases.

\section{Ablation}

\begin{table*}[ht!]
    \caption{Comparison of OOD detection results across different fusion types for Far-OOD datasets. Row highlighted in volor represent performances without logit fusion. Best and second best results are reported in bold and underline respectively.}
    \centering
    \label{tab:5}
    \resizebox{0.9\textwidth}{!}{%
    \begin{tabular}{lcccccccc}
        \toprule
        \multirow{3}{*}{\bf Fusion Type} 
        & \multicolumn{8}{c}{\bf OOD Dataset} \\
        \cmidrule{3-9}
        & \multicolumn{2}{c}{\bf OpenImage-O}  
        & \multicolumn{2}{c}{\bf NINCO}
        & \multicolumn{2}{c}{\bf Imagenet-O} 
        & \multicolumn{2}{c}{\bf Average} \\

        & AUROC $\uparrow$ &FPR95 $\downarrow$ 
        & AUROC $\uparrow$ & FPR95 $\downarrow$ 
        & AUROC $\uparrow$ & FPR95 $\downarrow$ 
        & AUROC $\uparrow$ & FPR95 $\downarrow$ \\
        
        \midrule
\rowcolor{blue!20}
        $S_{00}$ & $92.34\scriptstyle{\pm2.08}$ & $28.65\scriptstyle{\pm}3.44$& $73.96\scriptstyle{\pm1.03}$&$79.12\scriptstyle{\pm2.73}$ &$80.36\scriptstyle{\pm0.87}$ & $68.54\scriptstyle{\pm1.91}$&$82.22\scriptstyle{\pm1.32}$ &$58.77\scriptstyle{\pm2.69} $\\
\rowcolor{blue!20}
        $S_{aa}$ & $91.13\scriptstyle{\pm2.37}$ & $43.71\scriptstyle{\pm3.90}$ & $71.69\scriptstyle{\pm2.13}$ & $80.37\scriptstyle{\pm0.98}$ & $79.82\scriptstyle{\pm1.78}$ & $71.25\scriptstyle{\pm1.34}$ &$80.88\scriptstyle{\pm2.09}$ & $65.11\scriptstyle{\pm2.07}$\\

        \midrule

        $S_{00}+S_{aa}$ & $90.70\scriptstyle{\pm 2.31}$ & $46.90\scriptstyle{\pm 1.91}$ & $72.06\scriptstyle{\pm 3.49}$ &$81.61\scriptstyle{\pm 0.68}$ & $79.29\scriptstyle{\pm 2.70}$ & $77.55\scriptstyle{\pm 1.03}$ & $80.68\scriptstyle{\pm 2.83}$ & $68.69\scriptstyle{\pm 1.20}$ \\

        $S_{0a}+S_{aa}$ & $95.49\scriptstyle{\pm 0.52}$ & $19.33\scriptstyle{\pm2.19}$ 
        & $81.53 \scriptstyle{\pm 1.50}$ & $62.32\scriptstyle{\pm2.68}$
        & $86.87\scriptstyle{\pm 1.58}$ & $50.15\scriptstyle{\pm4.49}$ 
        & $87.96\scriptstyle{\pm 1.20}$ & $43.93\scriptstyle{\pm}3.12$ \\
        
        $S_{a0}+S_{aa}$ & $92.06\scriptstyle{\pm 0.85}$ & $38.95\scriptstyle{\pm3.77}$ 
        & $74.92\scriptstyle{\pm 1.82}$ & $79.23\scriptstyle{\pm3.81}$
        & $79.08\scriptstyle{\pm 4.95}$ & $74.45\scriptstyle{\pm3.38}$ 
        & $82.02\scriptstyle{\pm 2.54}$ & $64.21\scriptstyle{\pm}3.65$ \\

        $S_{0a}+S_{aa}+S_{a0}$ & $\underline{95.62\scriptstyle{\pm 0.11}}$ & $\underline{18.91\scriptstyle{\pm0.13}}$ 
        & $\underline{82.99\scriptstyle{\pm 0.73}}$ & $\underline{59.16\scriptstyle{\pm0.62}}$
        & $\underline{88.02\scriptstyle{\pm 0.51}}$ & $\underline{46.78\scriptstyle{\pm 0.69}}$ 
        & $\underline{88.87\scriptstyle{\pm 0.45}}$ & $\underline{41.61\scriptstyle{\pm0.48}}$ \\
        
        $S_{0a}+S_{aa}+S_{a0}+S_{00}$ & $\mathbf{96.07 \scriptstyle{\pm 0.68}}$ & $\mathbf{17.66 \scriptstyle{\pm 3.33}}$ 
        & $\mathbf{83.39 \scriptstyle{\pm 0.50}}$ & $\mathbf{59.38 \scriptstyle{\pm 2.19}}$ 
        & $\mathbf{87.77 \scriptstyle{\pm 2.30}}$ & $\mathbf{46.93 \scriptstyle {\pm 6.24}}$ 
        & $\mathbf{89.07\scriptstyle{\pm 1.16}}$ & $\mathbf{41.32\scriptstyle{\pm 3.92}}$ \\

        \bottomrule
    \end{tabular}
    }
\end{table*}

\begin{figure*}[ht!]
\centering
\pgfplotsset{
    shotsaxis/.style={
        width=\linewidth, height=0.55\linewidth,
        symbolic x coords={1,4,8,16,32}, xtick=data, xlabel={Shots},
        xlabel style={font=\small}, tick label style={font=\footnotesize},
        ylabel style={font=\small}, grid=major, major grid style={gray!25},
        line width=0.9pt, mark size=2.2pt, enlarge x limits=0.10,
    },
    vitB16/.style={color=blue!70!black, mark=*, mark options={fill=blue!70!black}},
    vitB32/.style={color=orange!90!black, mark=*, dashed, mark options={solid, fill=orange!90!black}},
    band16/.style={blue!70!black, opacity=0.15},
    band32/.style={orange!90!black, opacity=0.15},
}

{\centering
\begin{tikzpicture}[baseline]
  \draw[blue!70!black, line width=0.9pt] (-3.05,0) -- (-2.45,0)
        node[midway, fill=blue!70!black, circle, inner sep=1.6pt]{};
  \node[anchor=west, font=\footnotesize] at (-2.35,0) {ViT-B/16};
  \draw[orange!90!black, line width=0.9pt, dashed] (0.35,0) -- (0.95,0)
        node[midway, fill=orange!90!black, circle, inner sep=1.6pt]{};
  \node[anchor=west, font=\footnotesize] at (1.05,0) {ViT-B/32};
\end{tikzpicture}
\par}

\vspace{-0.4em}

\begin{subfigure}{0.31\textwidth}\centering
\begin{tikzpicture}
\begin{axis}[shotsaxis, ymin=85, ymax=100, ytick={85,90,95,100}, ylabel={AUROC}]
\addplot[draw=none, forget plot, name path=b16u] coordinates {(1,96.0)(4,91.8)(8,92.2)(16,96.0)(32,91.2)};
\addplot[draw=none, forget plot, name path=b16l] coordinates {(1,94.0)(4,89.2)(8,89.8)(16,94.0)(32,88.8)};
\addplot[band16, forget plot] fill between[of=b16u and b16l];
\addplot[draw=none, forget plot, name path=b32u] coordinates {(1,95.2)(4,91.3)(8,91.2)(16,94.2)(32,90.7)};
\addplot[draw=none, forget plot, name path=b32l] coordinates {(1,92.8)(4,88.7)(8,88.8)(16,91.8)(32,88.3)};
\addplot[band32, forget plot] fill between[of=b32u and b32l];
\addplot[vitB16] coordinates {(1,95.0)(4,90.5)(8,91.0)(16,95.0)(32,90.0)};
\addplot[vitB32] coordinates {(1,94.0)(4,90.0)(8,90.0)(16,93.0)(32,89.5)};
\end{axis}\end{tikzpicture}\caption{AUROC}\end{subfigure}\hfill
\begin{subfigure}{0.31\textwidth}\centering
\begin{tikzpicture}
\begin{axis}[shotsaxis, ymin=0, ymax=60, ytick={0,20,40,60}, ylabel={FPR95}]
\addplot[draw=none, forget plot, name path=f16u] coordinates {(1,21.0)(4,48.0)(8,45.0)(16,24.0)(32,42.0)};
\addplot[draw=none, forget plot, name path=f16l] coordinates {(1,15.0)(4,42.0)(8,39.0)(16,16.0)(32,34.0)};
\addplot[band16, forget plot] fill between[of=f16u and f16l];
\addplot[draw=none, forget plot, name path=f32u] coordinates {(1,25.0)(4,50.0)(8,49.0)(16,32.0)(32,44.0)};
\addplot[draw=none, forget plot, name path=f32l] coordinates {(1,19.0)(4,44.0)(8,43.0)(16,24.0)(32,36.0)};
\addplot[band32, forget plot] fill between[of=f32u and f32l];
\addplot[vitB16] coordinates {(1,18.0)(4,45.0)(8,42.0)(16,20.0)(32,38.0)};
\addplot[vitB32] coordinates {(1,22.0)(4,47.0)(8,46.0)(16,28.0)(32,40.0)};
\end{axis}\end{tikzpicture}\caption{FPR}\end{subfigure}\hfill
\begin{subfigure}{0.31\textwidth}\centering
\begin{tikzpicture}
\begin{axis}[shotsaxis, ymin=60, ymax=75, ytick={60,65,70,75}, ylabel={ID Accuracy (\%)}]
\addplot[draw=none, forget plot, name path=a16u] coordinates {(1,69.3)(4,71.7)(8,72.6)(16,73.6)(32,74.0)};
\addplot[draw=none, forget plot, name path=a16l] coordinates {(1,67.7)(4,70.3)(8,71.4)(16,72.4)(32,73.0)};
\addplot[band16, forget plot] fill between[of=a16u and a16l];
\addplot[draw=none, forget plot, name path=a32u] coordinates {(1,67.3)(4,69.3)(8,69.8)(16,70.3)(32,70.8)};
\addplot[draw=none, forget plot, name path=a32l] coordinates {(1,65.7)(4,67.7)(8,68.2)(16,68.7)(32,69.2)};
\addplot[band32, forget plot] fill between[of=a32u and a32l];
\addplot[vitB16] coordinates {(1,68.5)(4,71.0)(8,72.0)(16,73.0)(32,73.5)};
\addplot[vitB32] coordinates {(1,66.5)(4,68.5)(8,69.0)(16,69.5)(32,70.0)};
\end{axis}\end{tikzpicture}\caption{ID Accuracy}\end{subfigure}

\caption{Performance vs.\ number of shots. Results are averaged over three seeds with shaded standard deviation.}
\label{fig:shots}
\end{figure*}
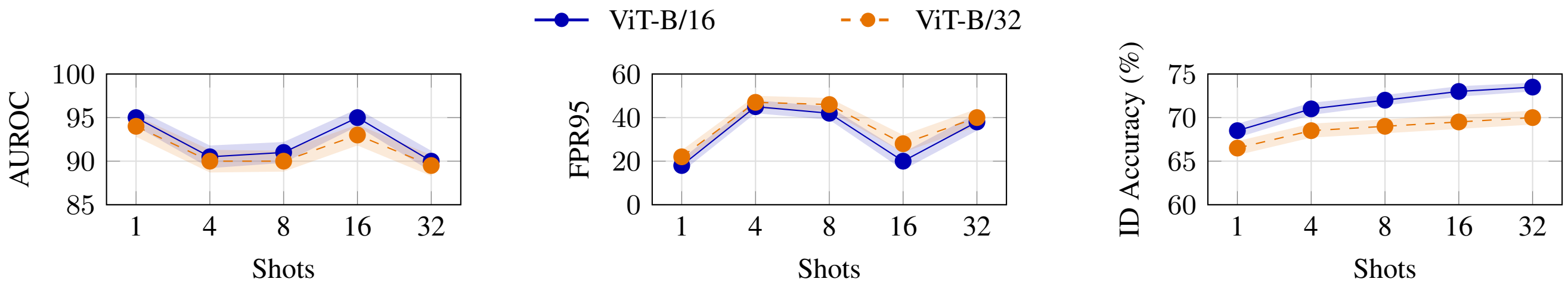

\subfour{Influence of Different Fusion Types:}
Table~\ref{tab:4} presents an ablation study analyzing the contribution of different similarity components in MM-Log for OOD detection. Here, we evaluate variants that progressively incorporate cross-modal interactions between zero-shot and adapted representations, namely $S_{00}+S_{aa}$, $S_{0a}+S_{aa}$, $S_{a0}+S_{aa}$, $S_{0a}+S_{aa}+S_{a0}$, and the full model $S_{0a}+S_{aa}+S_{a0}+S_{00}$. In addition to those, we also compare the purely zero-shot channel $S_{00}$, which uses the zero-shot image embedding and the zero-shot template text embedding with no trainable parameters, against its fully adapted counterpart $S_{aa}$, which uses only the two adapted branches. We can see that the methods without any similarity fusion (highlighted in color in Table~\ref{tab:4}) yield poor detection results. This highlights the shortcomings of the zero-shot-only method, and of methods that consider only the semantics of the downstream task.
 
\par Among the fusion types, the $S_{00}+S_{aa}$ variant performs the worst, with substantially degraded OOD detection performance. This behavior can be attributed to its reliance on similarities dominated by the adapted-only and zero-shot-only representations within each modality (image, and text). Without a proper cross-modal fusion formulation, the similarity scores can become more sensitive to downstream artifacts or overfit to generic CLIP embeddings from the zero-shot priors, leading to poor calibration and unreliable separation between ID and OOD samples. These findings highlight the importance of balanced similarity fusion to preserve robust semantics and ensure stable cross-modal similarity under distribution shift. Overall, this ablation confirms that effective OOD detection benefits from fusing complementary similarity signals, and that the proportionate combination of zero-shot priors and adapted interactions yields the most reliable performance.

\subfour{Performance Across Multiple-shots:}
In Fig. \ref{fig:shots}, we further analyze the effect of the number of training shots on both OOD detection performance and ID accuracy using two backbone architectures, ViT-B/16 and ViT-B/32. Here, we trained multiple models under different shot-settings: $\{1, 4, 8, 16, 32\}$, and then evaluated the average detection performance across Far-OOD benchmarks. Across both backbones, OOD detection performance as measured by AUROC and FPR95 exhibits a non-monotonic trend with respect to the number of shots: the strongest detection results are achieved in the low and mid-shot regimes, particularly at 1-shot and 16-shot settings. In contrast, increasing the number of shots consistently improves ID classification accuracy, with accuracy rising monotonically as more labeled data becomes available. Together, these results highlight a trade-off between detection and recognition: while additional shots benefit ID accuracy, they do not necessarily enhance OOD detection performance, and the 1-shot and 16-shot settings are most effective for detection. We attribute this non-monotonic behavior to a tension between the two objectives that the zero-shot priors only partially absorb. 

\section{Conclusion}
We presented MM-Log, a principled method for cross-modal OOD detection that explicitly addresses the miscalibration and brittleness of VLM similarity scores under distribution shift. By jointly leveraging zero-shot pretrained representations and task-adapted branches learned on downstream data, MM-Log strengthens alignment for ID image–text pairs while discouraging spurious or non-corresponding similarities that could mislead detection. Unlike prior methods, MM-Log does not require auxiliary OOD data, negative prompts, or web-scale label retrieval, yet it achieves consistently strong performance across Far-OOD, Near-OOD, and Hard-OOD benchmarks. Empirical results demonstrate that fusing zero-shot and adapted similarities yields better-calibrated OOD scores and robust ID–OOD separation. Our findings suggest that focusing on cross-modal alignment and embedding geometry, rather than relying on external OOD supervision, is an effective strategy for scalable and reliable OOD detection in large VLMs.

\bibliography{aaai2026}

\end{document}